# Sequence-to-Sequence Learning for Indonesian Automatic Question Generator

Ferdiant Joshua Muis
School of Electrical Engineering and Informatics
Institut Teknologi Bandung
Bandung, Indonesia
ferdiant.joshua@gmail.com

Ayu Purwarianti[1,2]
[1]U-CoE AI-VLB
[2]School of Electrical Engineering and Informatics
Institut Teknologi Bandung
Indonesia
ayu@informatika.org

*Abstract*—Automatic question generation is defined as the task of automating the creation of question given a various of textual data. Research in automatic question generator (AQG) has been conducted for more than 10 years, mainly focused on factoid question. In all these studies, the state-of-the-art is attained using sequence-to-sequence approach. However, AQG system for Indonesian has not ever been researched intensely. In this work we construct an Indonesian automatic question generator, adapting the architecture from some previous works [1][2]. In summary, we used sequence-to-sequence approach using BiGRU, BiLSTM, and Transformer with additional linguistic features, copy mechanism, and coverage mechanism. Since there is no public large and popular Indonesian dataset for question generation, we translated SQuAD v2.0 factoid question answering dataset, with additional Indonesian TyDiQA dev set for testing. The system achieved BLEU1, BLEU2, BLEU3, BLEU4, and ROUGE-L score at 38.35, 20.96, 10.68, 5.78, and 43.4 for SQuAD, and 40.01, 20.68, 10.28, 6.44, 44.17 for TyDiQA, respectively. We found that the system performed well when the expected answers are named entities and are syntactically close with the context explaining them. Additionally, from native Indonesian perspective, the best questions generated by our best models on their best cases are acceptable and reasonably useful.

*Keywords—AQG; sequence-to-sequence; factoid; BiGRU; BiLSTM; Transformer; linguistic features; copy mechanism; coverage mechanism; SQuAD v2.0; TyDiQA.*

## I. INTRODUCTION

In education, the task of creating question for evaluating students' understanding and performance is a very common work, and until now this task is done manually by the education workforces. However, the required number of questions is occasionally big, and some questions are created ineffectively repetitively. This task does consume lots of time and reduce the efficiency of the teacher [3]. Researchers have studied the process of evaluating student performance by generating questions and decided that the automation of question generation can significantly reduce the weights of teachers' task. Therefore, an effort to create an AQG system is started.

AQG applications have been researched and developed for more than 10 years [2] with various approaches, methods, and techniques. Some most common methods in AQG applications in the previous decade are cloze test, to the use of semantic pattern matching, syntactic features, and question template. These methods work using the conventional rule-based system. Nowadays, deep learning approach is much more common.

Cloze test method to evaluate student reading comprehension works by removing some words repetitively (for example one word every 5 words) and assigns the students to guess the missing word. This method has been used in classroom since 1950 [3]. Syntactic or semantic based parsing based parsing AQG methods works by parsing the reading passage by its syntactic or semantic meaning used in the language, then by generating the question with given hand-crafted rules [4]. The most common approach todays, deep learning, usually utilize sequence-to-sequence learning which consists of encoder and decoder. Within the last 4 years, AQG built with deep learning have improved considerably from years to years, keep achieving new state-of-the-arts.

Deep learning approaches have some significant strengths compared to the rule-based AQG. Although AQG with methods such as Cloze Test and rule-based are much easier to create, they rely heavily on hand-crafted rules. Therefore, with their ease of control, they mostly suit for small scale AQG, and become less and less efficient as the AQG domain expands. Deep learning does have better generalization compares to the rule-based ones. Despite the high complexity in developing deep learning system, if trained correctly the system will be able to construct its own rules in the process and scale up coherently with the size of the dataset.

Within the existing AQG system, none of those are developed for Indonesian. Therefore, in this work we are trying to implement an Indonesian AQG system using the architecture which is known to give the best result: sequence-to-sequence learning.

## II. RELATED WORKS

In summary, there are two kinds of AQG sequence-to-sequence architecture approaches. The first one uses RNN, either LSTM [5] or GRU [6] with their bidirectional [7] counterpart, and the second one uses Transformer [8]. All these works were built using SQuAD dataset, except for the first mentioned work below. Other than that, the evaluation schema across these studies were not very consistent among one another.

Serban, et al. [9] used knowledge graph as its first step, the creation of model input representation. The knowledge graph is constructed from the dataset, with question is associated with subject, fact is associated with relationship, and answer of the question is associated with the object. They used SimpleQuestion dataset which contains 100,000 question-answer pairs. They designed the sequence-to-sequence architecture to follow translation task architecture. The encoder is called a sub-encoder, which functions to retrieve three inputs which are called subject atom, relationship atom, and object atom, wrapped as one becoming fact embedding. The encoder was trained separately, before training the decoder. This fact embedding then become input for the decoder using GRU. This model achieved 33.32, 35.38 on BLEU, and METEOR, respectively.

Du, Shao, Cardie [1] developed its system by following Bahdanau, Cho, and Bengio architecture [10] with Luong, Pham, and Manning's attention [11]. Their AQG implemented two types of model, the first only uses a single encoder called sentence encoder, and the second uses two encoders which are sentence encoder and paragraph encoder. On the inference process, they used beam search with simplified copy mechanism to handle OOV words which replace UNK token with word from source sentence with the highest attention. This model achieved 17.50, 12.28, 16.62, 39.75 on BLEU3, BLEU4, METEOR, ROUGE-L.

Harrison and Walker [2] implemented their AQG using BiLSTM for encoder, and LSTM for decoder with Bahdanau's attention. The differentiating aspect of this model is the use of linguistic features: Named Entity (NE), Part-of-speech tag (POS), binary cased word indicator (case), and coreference resolution (CoRef). This system also uses See's copy mechanism [12]. This model achieved 19.98, 22.26, and 48.23 on BLEU4, METEOR, and ROUGE-L.

Kumar, Ramakrishnan, and Li [13] created a new framework in their AQG implementation. They used a novel framework called *generator-evaluator*. This framework works with *reinforcement learning* principle, mainly optimizes backward propagation and weight update by defining a new loss function. The generator is a sequence-to-sequence architecture like Bahdanau's with additional answer location, and linguistic features. The generator also uses copy and coveraged mechanism. The evaluator is the reward provider based on two reward functions called *Question Sentence Overlap Score* (QSS), and *Predicted and Encoded Answer Overlap Score* (ANSS). This model achieved 22.01, 16.48, 20.21, and 44.11 on BLEU3, BLEU4, METEOR, and ROUGE-L.

Liu, et al, [14] implemented their AQG system by using a new *clue indicator* which helps question generation. The clue indicator is generated by training a *Graph Convolutional Network* (GCN). This clue, along with its additional features become the input for the BiLSTM-LSTM sequence-to-sequence architecture. This model achieved 22.82, 17.55, 21.24 and 44.53 on BLEU3, BLEU4, METEOR, and ROUGE-L.

Dong, et al [15] AQG system combines four types of language model (LM) including Transformer. This model receives word-embedded token, token position, and sentence segment. Those inputs are then processed by using *parameter-sharing-transformer*, and the output is used to build the four LM with three types of mechanism attention. This model achieved 23.75, 25.61, and 52.04 on BLEU4, METEOR, and ROUGE-L.

## III. OUR PROPOSED SYSTEM

Most of the previous studies stated that AQG models developed with deep learning approach heavily outperform their rule-based counterpart, with higher flexibility and larger domain scope too. Acknowledging this phenomenon, we propose an Indonesian AQG using sequence to sequence approach. Generally, we designed two main models, RNN-based, and Transformer-based. We used paragraph context sentences along with some additional linguistic features: answer location (*ans*), case indicator (*case*), part of speech (POS), and named entity (NE) as the input, and question sentences as the target. For the dataset, as there is no large Indonesian question-answering dataset, we started by preparing the dataset, using translated SQuAD v2.0. Then we applied a correction preprocessing step to the dataset as predefined SQuAD's answer position need to be adjusted after translation. The overview of this AQG system building flow can be seen in Fig. 1.

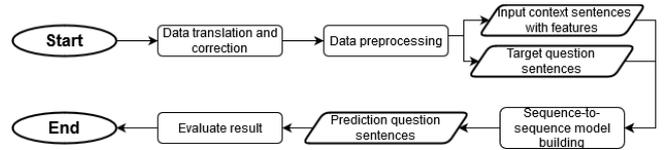

Fig. 1. AQG system building flow

### A. Data Translation and Correction

We used Google Translate API v2 to translate the SQuAD v2.0 from English to Indonesian for about 536 articles with 161.550 question-answer pairs. Since the data is quite large, we used the translation result as it is and there were no translation corrections applied. Instead, the dataset post-processing action mainly focuses on updating the answer and answer location value.

TABLE I.  SQUAD V2.0 TRANSLATION RESULT EXAMPLES

| Passage |
|---|
| 'context': '*Beyoncé Giselle Knowles-Carter (/ biːˈjɒnseɪ / bee-YON-say) (lahir 4 September 1981) adalah penyanyi, penulis lagu, produser dan aktris rekaman Amerika. Dilahirkan dan dibesarkan di* Houston, Texas, *ia tampil di berbagai kompetisi menyanyi dan menari sebagai … Hiatus mereka melihat rilis album debut Beyoncé,* Dangerously in Love *(2003), yang ...*'} |
| **Question-answer Pair 1** |
| {'question': '*Di kota dan negara bagian mana Beyonce tumbuh?*', 'id': '56bf6b0f3aeaaa14008c9601', 'answers': [{'text': 'Houston, Texas', 'answer_start': 166}], 'is_impossible': False} |
| **Question-answer Pair 2** |
| {'question': '*Apa nama album solo pertama Beyoncé?*', 'id': '56d43ce42ccc5a1400d830b5', 'answers': [{'text': '*Berbahaya dalam Cinta*', 'answer_start': 505}], 'is_impossible': False} |

The raw result of translated dataset was not usable directly with some problems showing up: (1) answer location mismatch, and (2) inconsistent translation result. SQuAD dataset stores its answers by using 2 values, the answer text themselves, and their character-based locations in the paragraph. Since the context paragraph text and answer text were translated separately, the answers sometimes had different translation result, which example is shown in Table I (2[nd] pair) where the named entity "Dangerously in Love" was translated into "Berbahaya dalam Cinta" in the answer text. This prevented us to find the location of the answer in the context paragraph text using exact matching. Thus, we conducted fuzzy string matching to repair the answer. The process is illustrated in Fig. 2. We used fuzzy string-matching search to enable the system to match inconsistent translation results, which usually still roughly match more than 80% using Levenshtein edit distance method. After this answer translation correction process, each data contained 2 new keys, *indonesian_answer_start* and *indonesian_answer*. The *indonesian_answer_start* key holds the new answer location or -1 if not found; the *indonesian_answer* key holds the new answer text or empty string if not found.

### B. Training-Testing Data Preparation

This process aims to adjust the translated dataset to be able to be used in the model training and testing phase. It consists of 10 steps:

*1) Discard every question-answer pairs which previous preprocessing step unable to find*: Due to the imperfection of the algorithm, some answers were still not found in the passage. These problematic question-answer pairs were marked with -1 value for the *indonesian_answer_loc* key, and then removed from the dataset.

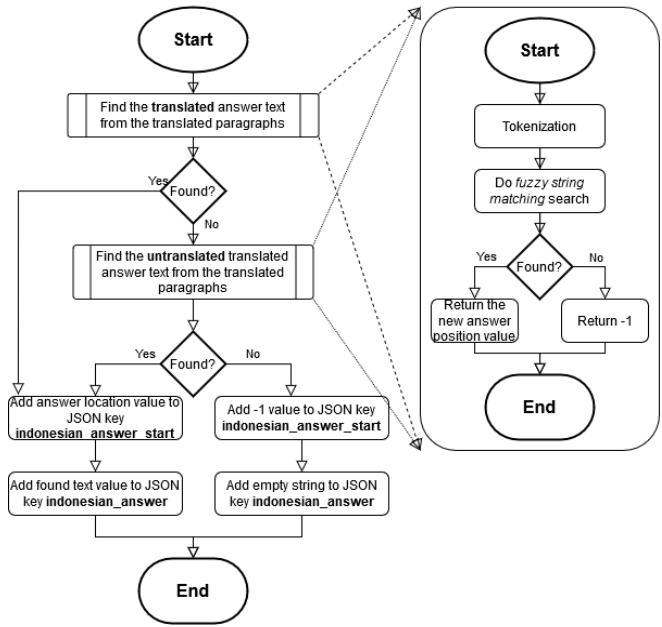

Fig. 2. Process to Repair Answer Translation Result

*2) POS and NE feature preparation*: We used third party Prosa.ai's POS Tagger v1.0 and Named Entities Extractor v1.0 API which categorize the POS and NE into 29 POS tags, and 19 NE tags, respectively. The intuition of using these features is because human will nearly always consider the POS and NE of the source passage while generating the questions.

*3) Remove outliers*: To simplify and accelerate model building, we removed very long question-answer pairs. We try to keep as many data as possible, so we removed data which contain the first 1% longest question or answer. As the result, the data contain questions with maximum of 60 words, and answers with maximum of 20 words.

*4) Word normalization and tokenization*: We converted every unicode characters to their closest ascii representations, then we splitted the text by space, and punctuations. Additionaly, we did not split by ".", ",", and ":" which do not follow and preceed space to handle some special cases such as time (e.g. "23:00"), large numbers (e.g. 20,000), degree (e.g. "Ph.D"), etc.

*5) Convert answer location from character-based to word-based*: As the data was tokenized and word-based, we converted the answer location accordingly.

*6) Create paragraph-question pairs based on word-level answer location*: We used context paragraph text as input, and question text as target. Every question has its corresponding pair sentence in the context paragraph text. We only took one sentence which contains the first word of the answer for the corresponding question. We did not handle questions which answers spreaded in more than one sentence specifically.

[1]https://prosa.ai
[2]https://github.com/facebookresearch/fastText

*7) Create linguistic features*: We used 4 features to accompany the model input: answer location (ans), case indicator (case), part of speech (POS), and named entity (NE). For both ans and case feature, we created additional binary-valued arrays having the same length as the input. Any elements were switched from 0 to 1 if the corresponding input sentences' word (from step 6) in the same position contains answer or cased letters. For the POS and NE, we aligned the results from Prosa.ai's POS Tag and Named Entity Extractor API with the input sentencse.

*8) Synchronize and align POS and NE features with the input*: There were some little differences between the tokenization method used by our implementation and Prosa.ai[1] API's implementation. Therefore, we aligned them word by word, before converting them to integer representation.

*9) Create uncased input variation*: the experiment consisted of cased and uncased input variations.

*10) Create integer representations for each input, features, and target, as well as weight matrix from FastText[2] pretrained word-embedding*: This step was only applied to the self-implementation model which we will discuss next.

These steps leave the amount of question-answer pairs data down to 114,604 and 10,536 for training and testing dataset, respectively. We splitted the data by 9:1 for training-validation set, resulting in 102,657 training data and 11,407 validation data. The training and validation dataset originate from SQuAD training set, and the testing dataset originates from SQuAD dev set.

### C. Sequence-to-Sequence Model

We designed the architecture into two primary types: RNN and Transformers, with BiGRU and BiLSTM topologies were used as the RNN, and implemented either Bahdanau et al.'s [10] or Luong et al.'s [11] attention mechanism. Each of these three algorithms was then extended to have cased/uncased, copy mechanism, coverage mechanism variations. We didn't implement coverage variation for Transformers, as we implemented See et al.'s copy and coverage mechanism [12], and See et al.'s coverage mechanism does not suit with Transformer's attention mechanism.

We adapted the RNN architecture from Du et al.'s [1] with some modifications: replacement of GloVe's word embedding with FastText's, additional linguistic features (*ans*, *case*, POS, NE), and the absence of sentence embedding encoder. For the Transformer architecture, we adapted the implementation from Vaswani's [8] with additional linguistic features as RNN's. Both of RNN and Transformer architecture can be seen in fig. 3 and fig. 4, respectively.

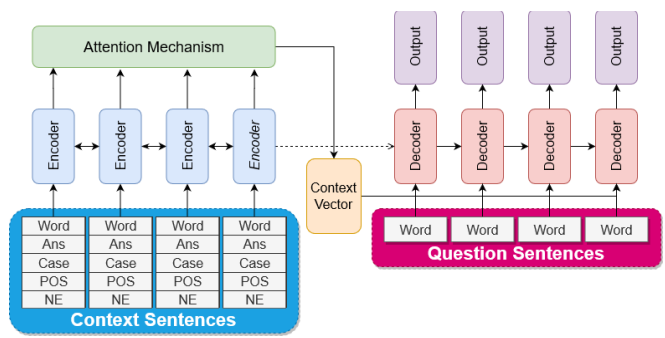

Fig. 3. RNN architecture for Our Indonesian AQG

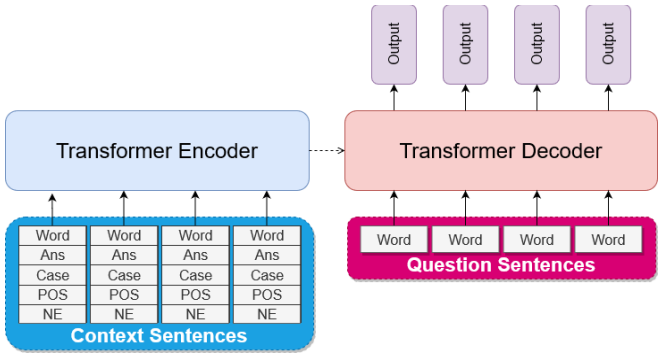

Fig. 4. Transformer architecture for Our Indonesian AQG

*D. Evaluation*

Most of AQG studies use BLEU, METEOR, and ROUGE-L. In this research, we omitted the METEOR from our evaluation scenarios. This is due to METEOR also takes into account stems, synonyms, and phrases to calculate its score, and this indeed cause the metric to be language dependent. Currently there is no well-concerned studies for Indonesian METEOR, therefore we decided to only use 5 metrics: BLEU-1, 2, 3, 4, and ROUGE-L. We used Sharma et al.'s NLG evaluation library [16], NLGEval. Although these metrics are usually used in machine translation task, we considered they are feasible to be used in QG evaluation. As these metrics assess score based on prediction and target similarity (ranging from 0-1 but we multiply them by 100 to simplify reading and comparison), this match the aim of evaluating generated questions in supervised manner; even though we cannot do much about the missed semantic meaning evaluation.

We noticed that the imperfection of SQuAD translation creates lots of unnatural Indonesian sentences. There was a considerable amount of translation errors, syntactically and semantically, which do not share a common pattern, requiring mass of human efforts to fix. Therefore, in addition to evaluating on SQuAD's development set, we also evaluate on Indonesian TyDiQA's development set. Every data in TyDiQA's dataset was collected manually from texts written in the original language, resulting in much more natural sentences, including Indonesian. We hope that by also evaluating on more natural Indonesian dataset, we can obtain much more reliable and comparable results.

## IV. EXPERIMENTS

This part covers the experiment implementations[3] following the prementioned proposed system. These experiments had two major goals: (1) to build AQG system using various sequence-to-sequence approaches, and (2) to measure and compare the performance of each models. We used 300-dimension Indonesian FastText for the word embedding and split the training-validation set with 9:1 ratio. Eventually, we ended up with 102,657 training data, 11,407 validation data, 10,567 testing data for SQuAD, and 550 testing data for TyDiQA. We implemented the experiments using PyTorch [17] in two different methods: (1) self-implementation, and (2) OpenNMT-py[4] library [18] – which is also based on PyTorch. Hence, we categorized the conducted experiments into two scenarios with different intentions: (1) to find out which method either self-implementation of OpenNMT performs better, and (2) to build the most performing AQG model. It was later found that OpenNMT implementation outperformed self-implementation, consequently the second scenario was fully implemented using OpenNMT library. All experiments were run on a single core NVIDIA V100 GPU.

*A. RNN-Based (BiGRU & BiLSTM)*

We trained BiGRU and BiLSTM on as-similar-as-possible hyperparameters to allow us to compare the models fairly.

We trained 2 different types of model for the first scenario: the first was self-implemented (named as BiLSTM-1 & BiGRU-1), and the second was using OpenNMT (BiLSTM-2 & BiGRU-2). On the second scenario, we only trained 1 type of model, but the model was variated with copy mechanism, coverage mechanism, as well as the cased/uncased variation, named as BiLSTM-3 & BiGRU-3, with appended "-Copy", "-Coverage", "-Cased", and "-Uncased" label corresponding to the used variations. In short, for the first scenario we trained and compared BiGRU-1, BiGRU-2, BiLSTM-1, and BiLSTM-2. For the second scenario, we trained and compared BiGRU-3 & BiLSTM-3 along with their variations.

*B. Transformer-Based*

The Transformer-based experiments shared same concept as the RNN-based's: two same scenarios, same variations, same objectives, and same adapted hyperparameters from previous works. Some differences are, we adapted the hyperparameters from Vaswani et al.'s Transformer, and these models do not use coverage variation, as our coverage mechanism (See et al.'s) does not support Vaswani et al.'s attention model. Like RNN-based model experiments, for the first scenario we trained and compared Transformer-1 and Transformer-2. And for the second scenario, we trained and compared Transformer-3 along with its variations.

## V. RESULTS AND ANALYSIS

*A. Performance Comparison of Self-implementation and OpenNMT Implementation Scenario*

Our experiments showed that OpenNMT implementation greatly outperformed the self-implementation. On all metrics, OpenNMT implementation had much greater score, except for Transformer's training time which takes 2.66 times slower than our self-implementation. The complete result can be seen in Table II. Some models have different training epoch between one another, depends on when the models stopped learning and started overfitting. Based on this finding, we chose to optimize the model by using OpenNMT implementation in the second scenario.

TABLE II. PERFORMANCE COMPARISON OF SELF-IMPLEMENTATION MODELS AND OPENNMT IMPLEMENTATION MODELS

| Model | BLEU | | | | ROUGE L | *Epoch* | Training Time (m) |
|---|---|---|---|---|---|---|---|
| | 1 | 2 | 3 | 4 | | | |
| BiGRU-1 | 20.19 | 4.94 | 0.83 | 0.16 | 23.44 | 15 | 49 |
| BiGRU-2 | **29.43** | **12.86** | **5.26** | **2.39** | **35.69** | 20 | **32** |
| BiLSTM-1 | 20.43 | 5.17 | 0.89 | 0.21 | 23.72 | 15 | 50 |
| BiLSTM-2 | **22.93** | **6.23** | **1.19** | **0.21** | **27.87** | 20 | **31** |
| Transformr-1 | 22.48 | 5.8 | 12.2 | 0.28 | 25.9 | 70 | **117** |
| Transformr-2 | **29.06** | **11.66** | **4.37** | **1.67** | **33.48** | 70 | 310 |

*B. Model Optimization using OpenNMT Scenario*

Performance comparison between our most optimized models can be seen in Table III for SQuAD test set and Table IV for TyDiQA test set. Replacement of unknown token ("<unk>") was involved in the inferencing step, preventing the model from generating unknown token. This unknown token replacement was done by replacing the token with source token which has the highest attention value.

---
[3] The code is available at https://github.com/FerdiantJoshua/question-generator
[4] https://github.com/OpenNMT/OpenNMT-py

We found that uncasing the input sentences did improve model performance. All uncased variants achieved higher metric compared to the cased ones. Copy and coverage mechanism were also found to be effective in increasing higher model quality, with copy mechanism had much more significant effect than coverage mechanism. Overall, the most performing model was BiGRU-3-Uncased-Copy-Coverage, followed by BiLSTM-3-Uncased-Copy-Coverage, and finally Transformer-3-Uncased-Copy.

Models that were trained using copy mechanism in uncased variant were seen to be the ones which can produce higher metric result when tested on TyDiQA test set, compared with when tested on SQuAD test set. This phenomenon was found on every models. We analyzed the reason why this occurred is due to the smaller size of possible word choices on uncased variation, and the ability to copy from the source sentences. Smaller choice to make means more contextual prediction, and an ability to copy makes it much easier to ask by copying part of primary phrases or named entities as parts of the question. Moreover, by uncasing the input sentences, two identical words with a single difference in capital letter will have same word embedding in the model (e.g. "duchy" and "Duchy").

TABLE III. PERFORMANCE COMPARISON OF GRU, LSTM, AND TRANSFORMER MODELS ON SQUAD

| Model | BLEU | | | | ROUGE L | Epoch |
|---|---|---|---|---|---|---|
| | 1 | 2 | 3 | 4 | | |
| **BiGRU-3** | | | | | | |
| Cased | 32.59 | 15.09 | 6.73 | 3.35 | 36.82 | 20 |
| Cased-Copy | 35.75 | 18.74 | 9.61 | 5.3 | 40.28 | 20 |
| Cased-Copy-Coverage | 35.63 | 18.67 | 9.51 | 5.22 | 40.25 | 20 |
| Uncased | 35.42 | 17.1 | 7.37 | 3.55 | 39.32 | 20 |
| Uncased-Copy | 38.66 | 21.03 | 10.68 | **5.87** | 43.02 | 20 |
| Uncased-Copy-Coverage | **39.2** | **21.48** | **10.88** | 5.86 | **43.32** | 20 |
| **BiLSTM-3** | | | | | | |
| Cased | 31.75 | 14.17 | 6.03 | 2.84 | 36.08 | 10 |
| Cased-Copy | 35.44 | 18.43 | 9.33 | 5.11 | 40.02 | 10 |
| Cased-Copy-Coverage | 35.48 | 18.47 | 9.29 | 5.08 | 40.06 | 10 |
| Uncased | 35.08 | 16.56 | 6.91 | 3.22 | 38.97 | 10 |
| Uncased-Copy | 38.92 | 21.09 | 10.69 | 5.75 | **43.1** | 10 |
| Uncased-Copy-Coverage | **39.14** | **21.27** | **10.72** | **5.93** | 43.08 | 10 |
| **Transformer-3** | | | | | | |
| Cased | 29.33 | 11.33 | 3.96 | 1.35 | 33.38 | 300 |
| Cased-Copy | 34.4 | 17.81 | 8.92 | 4.64 | 39.05 | 300 |
| Uncased | 32.03 | 13.59 | 4.7 | 1.72 | 35.91 | 300 |
| Uncased-Copy | **38.14** | **20.57** | **10.16** | **5.38** | **42.37** | 300 |

TABLE IV. PERFORMANCE COMPARISON OF GRU, LSTM, AND TRANSFORMER MODELS ON TYDIQA

| Model | BLEU | | | | ROUGE L | Epoch |
|---|---|---|---|---|---|---|
| | 1 | 2 | 3 | 4 | | |
| **BiGRU-3** | | | | | | |
| Cased | 31.41 | 12.2 | 4.18 | 1.96 | 35.34 | 20 |
| Cased-Copy | 35.35 | 16.15 | 8.08 | 4.51 | 39.35 | 20 |
| Cased-Copy-Coverage | 35.53 | 15.69 | 7.7 | 4.53 | 39.46 | 20 |
| Uncased | 33.89 | 14.0 | 5.38 | 2.88 | 37.98 | 20 |
| Uncased-Copy | 39.12 | 19.87 | **10.36** | 6.24 | 43.39 | 20 |
| Uncased-Copy-Coverage | **40.01** | **20.68** | 10.28 | **6.44** | **44.17** | 20 |
| **BiLSTM-3** | | | | | | |
| Cased | 31.0 | 11.95 | 3.89 | 1.37 | 35.12 | 10 |
| Cased-Copy | 35.0 | 15.13 | 7.04 | 4.16 | 38.97 | 10 |
| Cased-Copy-Coverage | 35.06 | 15.68 | 7.72 | 4.6 | 39.12 | 10 |
| Uncased | 33.21 | 13.66 | 5.04 | 2.53 | 37.39 | 10 |
| Uncased-Copy | 39.39 | **19.71** | 10.11 | **6.37** | 43.67 | 10 |
| Uncased-Copy-Coverage | **39.7** | 19.63 | **10.14** | 6.23 | **43.86** | 10 |
| **Transformer-3** | | | | | | |
| Cased | 27.51 | 8.24 | 0.95 | 0.31 | 31.49 | 300 |
| Cased-Copy | 33.8 | 13.9 | 5.7 | 3.16 | 38.06 | 300 |
| Uncased | 30.68 | 10.82 | 2.88 | 1.0 | 34.77 | 300 |
| Uncased-Copy | **38.69** | **19.42** | **9.0** | **5.04** | **43.36** | 300 |

TABLE V. GENERATED QUESTION EXAMPLES FROM SQUAD TEST SET

| 1 | |
|---|---|
| Input Sentence & Answer | Kadipaten Normandia , yang mereka bentuk dengan perjanjian dengan mahkota Prancis , adalah tanah yang indah bagi Prancis abad pertengahan , dan di bawah **Richard I** dari Normandia ditempa menjadi sebuah pemerintahan yang kohesif dan tangguh dalam masa jabatan feodal . |
| Target Qstn | Siapa yang memerintah kadipaten Normandia |
| **BiGRU-3** | |
| Uncased | kekaisaran mana yang menjadi penguasa kekaisaran ? |
| UC-Copy | siapa yang memimpin kadipaten normandia ? |
| UC-Cop-Cov | siapa yang memerintah pemerintahan normandia ? |
| **BiLSTM-3** | |
| Uncased | siapa yang menguasai kadipaten normandia ? |
| UC-Copy | siapa yang memerintah normandia ? |
| UC-Cop-Cov | siapa yang mendirikan kadipaten normandia ? |
| **Transformer-3** | |
| Uncased | siapa yang menjadi penguasa hongaria ? |
| UC-Copy | siapa yang memerintah normandia ? |
| **2** | |
| Input Sentence & Answer | Gaya konservatif yang bekerja pada sistem tertutup memiliki kerja mekanis terkait yang memungkinkan energi untuk mengkonversi hanya antara bentuk **kinetik** atau potensial . |
| Target Qstn | Apa bentuk energi potensial yang dapat berubah ? |
| **BiGRU-3** | |
| Uncased | apa bentuk yang memungkinkan energi untuk mengubah energi ? |
| UC-Copy | apa yang memiliki kerja mekanis ? |
| UC-Cop-Cov | apa yang memiliki kerja mekanis untuk mengkonversi energi ? |
| **BiLSTM-3** | |
| Uncased | gaya konservatif apa yang digunakan oleh gaya konservatif ? |
| UC-Copy | gaya konservatif yang bekerja pada sistem tertutup memiliki bentuk apa ? |
| UC-Cop-Cov | gaya konservatif yang memiliki energi mekanis terkait terkait bentuk apa ? |
| **Transformer-3** | |
| Uncased | jenis energi apa yang digunakan sistem kerja ? |
| UC-Copy | gaya konservatif yang bekerja pada sistem tertutup antara bentuk apa ? |
| **3** | |
| Input Sentence & Answer | Celoron terus ke selatan sampai ekspedisinya mencapai pertemuan di Ohio dan sungai Miami , yang terletak tepat di sebelah selatan **desa Pickawillany** , rumah kepala suku Miami yang dikenal sebagai " Orang Inggris Tua " . |
| Target Qstn | Di mana Old Briton tidak menelepon ke rumah ? |
| **BiGRU-3** | |
| Uncased | di mana miami berada ? |
| UC-Copy | desa apa yang dikenal sebagai " orang inggris tua " ? |
| UC-Cop-Cov | di mana sungai miami berada ? |
| **BiLSTM-3** | |
| Uncased | di desa apa miami berada ? |
| UC-Copy | di mana rumah kepala suku miami ? |
| UC-Cop-Cov | di mana rumah kepala suku miami ? |
| **Transformer-3** | |
| Uncased | di mana miami berada ? |

| UC-Copy | *di mana celoron berada ?* |

Table V shows some generated question from each of our models on a sample of SQuAD test set. "Input sentence" is the model input, "Answer" is the expected answer, which is also used in *ans* feature, and "Target Qstn" is the expected generated question. Some model names are shortened to fit in the table. Overall, there were found lots of imperfect and unnatural translations on the input sentences and target questions, indeed affecting our model prediction. However, those sentences were still able to be understood semantically. In three cases showed in Table V, we analyze the result as: (1) First case: All models were found to agree on the same predicted questions, resulting very similar questions. There were also some differences on the used verb in the generated questions, but they all are synonyms and have similar meaning. (2) Second case: No models generated any questions good enough to semantically match with the target questions, or at least the expected answer, despite the imperfect target question due to translation. We suspected that this happened as the expected answer is not a named entity, and the context explaining the answer is syntactically far away. (3) Third case: There were too many problems (mainly translation's) which caused none of the models predict any question which were similar enough with the target: a named entity, "Old Briton" got accidentally translated, and then phrase "didn't call home" got translated very literally, additionally this question is also marked "*is_impossible*" on the dataset. However, our best models, RNN with copy and coverage, were able to generate acceptable questions which have the same answer as the expected one regardless. On the other hand, this kind of problem lowered overall BLEU and ROUGE-L score.

Lastly, the questions generated by GRU and LSTM were not significantly too different, indicating that small metric difference does not really affect their performance quality. While Transformer, having the largest evaluation metric score difference, were found to have difficulty in understanding semantic context. But we suspected it happened as the Transformers had not yet been trained well enough to reach their lowest minima as low as their RNN counterparts.

VI. CONCLUSION AND FUTURE WORK

Our experiments have proved that building Indonesian AQG system using as-is machine-translated question answering dataset (SQuAD v2.0) is possible with acceptable result, although it makes the model to learn from biased and unnatural data, affecting its generation questions occasionally. We also found that implementing using OpenNMT does allow us to build AQG system much more effectively and efficiently, compared to implementing by ourselves. Furthermore, the model variation: Uncased/Cased, copy mechanism, and coverage mechanism does help improving model performance significantly. From native Indonesian perspective, we consider the questions generated by our best models on their best cases to be acceptable and reasonably useful.

In the future, it will be worth to explore, or prepare a more natural Indonesian dataset. This includes designing a better and more precise preprocessing methods to filter bias and syntactically incorrect data, or even creating a new Indonesian question-answering/generation dataset. We also suspect that our best model we have got in these experiments can be improved with a more precise hyperparameters. Lastly, we also recommend the construction of AQG system using the latest Transformer's variations which have been found to achieve a plentiful amount of state-of-the-art models in many other NLP tasks such as GPT2, GPT3, XLM, BART, T5, etc.

ACKNOWLEDGMENT

This research is partly funded by ITB P3MI research for Informatics research group with title of "Deep Learning Implementation with Sequence to Sequence Technique for Automatic Question Generator".